# DEM Registration and Error Analysis using ASCII values

**Suma Dawn, Vikas Saxena, Bhu Dev Sharma**

**Abstract-**
Digital Elevation Model (DEM), while providing a "bare earth" look, is heavily used in many applications including construction modeling, visualization, and GIS. Their registration techniques have not been explored much. Methods like Coarse-to-fine or pyramid making are common in DEM-to-image or DEM-to-map registration. Self-consistency measure is used to detect any change in terrain elevation and hence was used for DEM-to-DEM registration. But these methods apart from being time and complexity intensive, lack in error matrix evaluation. This paper gives a method of registration of DEMs using specified height values as control points by initially converting these DEMs to ASCII files. These control points may be found by two mannerisms - either by direct detection of appropriate height data in ASCII files or by edge matching along congruous quadrangle of the control point, followed by sub-graph matching. Error analysis for the same has also been done.

## I. Introduction

DEM (Digital Elevation Models) are data files that contain the elevation of the terrain over a specified area, usually at a fixed grid interval over the "Bare Earth" [42], [43]. It epitomizes the elevation attribute of the terrain in discreet form in a 3-D space of a surface [1]. The intervals between each of the grid points will always be referenced to some geographical coordinate system (usually either latitude-longitude or UTM coordinate systems). The denser the grid points, more the detailed information will be in the file.

Among the enumerable applications, the generic *applications of DEM* include - extracting terrain parameters, modeling water flow or mass movement (say, avalanches and landslides), creation of relief maps, rendering of 3D visualizations including flight planning, creation of physical models, rectification of aerial photography or satellite imagery, terrain analyses in geomorphology and physical geography, Geographic Information Systems (GIS), engineering and infrastructure design, Global Positioning Systems (GPS), line-of-sight analysis, precision farming and forestry, Intelligent Transportation Systems (ITS), Advanced Driver Assistance Systems (ADAS). DEM's are also utilized in support of the pre-planning and lay-out of corridor surveys, seismic line locations, construction activities, and environmental purposes [2], [14], [11], [12], [42], [46].

*Sources of DEMs*: The following three main sources from which DEMs are usually contrived: (1) Interferometric Synthetic Aperture Radar (InSAR), (2) Stereographic images using the correlation among digital images [8], (3) interpolating of digital contour or topographic maps [12]. For creation of DEMs it is important to have the *elevation data sources* of the earth at the point on represented in the DEM, a few of which are: (a) Real Time Kinematic GPS, (b) stereo photogrammetry, (c) Theodolite or total station, (d) Doppler radar, (e) Focus variation, and (f) Inertial surveys. Satellites that provide good input for producing DEMs include Geosats, ERS-1/2 (ESA), Landsat TM (Landsat 4 and 5) and ETM+ (Landsat 7), ALOS, PRISM, PALSAR and other SARs, AVNIR-2 among others [8],[9],[10],[12],[16],[17].

*Representations of DEMs:* DEMs may be represented as: (i) Raster or grid (RSG – Regular Square Grids) also called Raster DEM and (ii) Triangular Irregular Network – (TIN) model or Vector DEM [7], [14], [22]. The *characteristics of DEM* so generated depends on certain parameters which are enumerated as follows: (i) terrain roughness; (ii) sampling density (elevation data collection method); (iii) grid resolution or pixel size; (iv) interpolation algorithm; (v) vertical resolution; (vi) terrain analysis algorithm; [12].

DEM registration is technique to represent DEM in congregation with either another DEM or map or an image. This paper is an attempt to do DEM-DEM registration at its initial phase and then follow it up with generalization, i.e., DEM-to-topographic maps and DEM-to-image registration. Section II describes briefly the current methods for DEM registration and a few of their shortcomings. Proposed method for doing DEM-to-DEM registration is discussed in Section III followed by result analysis in Section IV. The paper is wound-up by conclusion and future work projection in Section V.

## II. Preliminaries

DEM registration is important as it allows for seamless integration of DEMs the same place which may be represented in different orientation or may be processed using different resolutions. There are only a handful of methods for DEM registration most of which are DEM-to-topographical map registration or DEM-to-Remote Sensed Images. DEM-to-DEM registration is elusive because of the difficulties of finding feature points or control points required for matching and evaluation and error analysis.

Sefercik [1] used DEMSHIFT software for superimposing 2 DEMs. AML (ARC Macro Language) – another software created by ESRI for generating end-user applications in ARC/INFO Workstation was described by Ali, M.A.J.M., Adnan, R., Tahir, N.M., Rahman, M.H.F., Yahya, Z., Samad, A.M [23]. Li Yao, He Chen [30] used OPENGL for reading the USGS Standard format DEM data and gave two ways of displaying the digital terrain's surface: (i) Triangle mesh and (ii) Constructing surface using NURBS curve.

Bambang Trisakti and Ita Carolita [2], proposed **pyramid-layer-making** for DEM generation and comparison between ASTER Stereo Data and SRTM DEMs which used Ground Control Points (GCPs) by combining XY coordinate point and height (Z) point. Accuracy evaluation was done by drawing vertical and horizontal transect lines along both the DEM images, and then comparing the height distribution of each transect lines.

Zhengxiao Tony Li and James Bethel [3] performed two coarse DEM alignments, vertical alignment and horizontal alignment as an initial step before the fine DEM registration. Thereafter, least squares matching method was used.

In the paper by LI Yong, WU Huayi, [4], a new technique of DEM extraction from LIDAR data based on **morphological gradient** was proposed.

Maire & Datcu [5] constructed the visualization dataset by registering optical image and DEM using an object-based description of large optical EO images followed by fusion of the extraction information into the DEM by interactively selecting and classifying among a set of user-thematic. Hosford, S., Baghdadi, N., Bourgine, B., Daniels, P., King, C. [6] described the fusion of airborne laser altimeter data and a stereo-radargrammetric DEM. In the literature by Pauline Audenino, Loi c Rognant, Jean-Marc Chassery, Jean-Guy Planes, [7] TIN – Triangular Irregular network Modelling is used to model the DEM from the terrain using Delaunay triangulation method. Top-down and bottom-up approaches for fusion were compared. DEM production and fusion of SAR images using area-based grey value matching was done by Xinwu Li, Huadong Guo, Changlin Wang, Zhen Li, Jingjuan Liao [8]. Another GCP extraction algorithm based on geocoding of radar images requiring precise orbit information is mentioned in Sang-Hoon Hong, Jung, H. S., Won, J. S., Hong-Gab Kim [18]. For the classification and mapping of land cover types and other environmental monitoring and planning purposes, Bucher, T., Lehmann, F. [26] proposed a combination of the High Resolution Stereo Camera - Airborne (HRSC-A) and the HyMap hyperspectral scanner covers for generating a very high spatial & spectral resolution data and Digital Elevation Model (DEM) for the resulting Above Ground. Gauss Markov random fields (GMRF) models and a Bayesian approach were used to filter DEM. SAR TM and DEM fusion was shown by Takeuchi, S. [31]. Lahoche, F., Herlin, I. [35] proposed a scheme for generating daily high resolution maps: mapping individual temporal values on the classification image. Similar DEM-to-image and DEM-to-map registration method were described in [27], [28], and [29] also.

In the paper by Howard Schultz, Edward M. Riseman, Frank R. Stolle, Dong-Min Woo [33] proposed a novel method – **self-consistency measure** for multiple DEM-to-DEM registration by detecting changes in the terrain elevation. They showed two-image stereo reconstruction and multi-view stereo using feature matching and texture matching. They also divided the 3D reconstruction algorithms into image space methods and object-space matching. The basic principle is to identify outliers (unreliable elevations estimates) and then compute a weighted average of all reliable points.

The **Coarse to Fine (CTF)** hierarchical operation using image pyramids (3-5 layers) was adopted into the DEM generation process by Junichi Takaku, Noriko Futamura, Tetsuji Iijima, Takeo Tadono, Masanobu Shimada, Ryosuke Shibasaki [10]. Similar approach for fusion of orthophoto image and DEM generated from PRISM was utilized by N. Futamura, J. Takaku, H. Suzuki, T. Iijima, T. Tadono, M. Matsuoka, M. Shimada, T. Igarashi, R. Shibasaki [19]. **Permanent Scatters** are used for fusion of low resolution DEMs by Ferretti, A., Monti-Guarnieri, A., Prati, C., Rocca [20]. **PU (Phase Unwrapping)** method was used by Allievi, J., Ferretti, A., Prati, C., Ratti, R., Rocca, F.,

[25] for Multi-interferogram and DEM fusion. Alessandro Ferretti, Claudio Prati, and Fabio Rocca [32] proposed method for removal of phase distortions and combining DEMs in **wavelet domain** by means of a weighted average.

DEM calibration for estimation of systematic height errors was performed by Wessel, B., Gruber, A., Gonzalez, J.H., Bachmann, M., Wendleder, A [13] in which they used least square adjustment approach along with calibration reference data incorporating tie-point concept with GCP. Error analysis and validation were done by Jie Li, Haifeng Huang, Diannong Liang [24] using Monte-Carlo simulation. RMS accuracies were checked in Belgued, Y., Rognant, L., Denise, L., Goze, S., Planes, J.G., [15]. Steven Bowe [36] alsp performed Error and accuracy measurements.

Most of these methods apart from being time and complexity intensive, lack in error matrix evaluation. Also none prove to be suitable for different resolution DEMs.

### III. Proposed method for DEM-to-DEM registration.

DEM Registration may be thought of a set steps including:

1. Procurement of DEMs – atleast one against which the quality can be measured so it must be of high quality – considered to be the reference DEM against which error measurement and analysis is also done.
2. Finding quality feature points to be matched.
3. Alignment of DEMs to be registered. This includes Orthorectification.
4. Fusion of DEMs based on their Control Point matching and resolution
5. Enhancement and output generation.

DEMs may be procured freely from some of the sources which provide DEMs like (i) GTOPO30 (30 arcsecond resolution, approx. 1 km),(ii) Advanced Spaceborne Thermal Emission and Reflection Radiometer (ASTER) instrument of the Terra satellite - 30 meter resolution, (iii) Shuttle Radar Topography Mission (SRTM) data – 3 arc-second resolution (around 90 meters), (iv) TerraSAR-X:a German Earth observation satellite. Though not all but some of these provide good and high resolution (upto 1 arc) DEMs which may be used as a set of reference DEM for registration and error evaluation purpose.

Steps followed are:

1. Convert the DEMs to ASCII format. This can be performed directly from the DEMs themselves by using a freeware viewer like 3DEM [16], as shown in Fig 1 that allows user to convert and save the DEM to ASCII format. This ASCII file is a text file containing height values. Free DEMs may be downloaded from a number of sources. The DEM used in this project was from [45].

2. These ASCII files are then compartmentalize these to manageable sizes as it would require a colossal sized RAM for processing. The breaking down process from parent ASCII file to the children ASCII files doesn't require much processing and is hence easier to handle.

3. Feature point finding. This step involves getting the initial set of points from the user. Currently this is a manual process, so all the points are entered by the user. These points may be directly marked from either the DEM viewer or from the viewer created to read and render the ASCII file. A few control points have been shown as marked in Fig. 2. These points are then stored in a file. The file stores 3 parameters for each point marked – the latitude, longitude, and the elevation data. These control points to be matched are then found in the candidate DEM. The experimentation presently done considers two methods of doing this. First, since, ASCII data is being used, usually the data matched is accurate, but, some times, due to the difference in resolution at which these DEMs have been processed, the data matched gives a few false positives. The second method consists of edge matching along congruent quadrangle of the control point.

4. From the ASCII file, with reference to the initial DEM, retrieve the latitude and longitude combination (the best match is considered) in the DEMs to be registered. Presently since the DEMs have either the latitude or longitude alignment, it is not too intricate to match.

5. Form a parent graph of the points considered as control points from the reference DEM. Once these points are found in one or multiple files, form graphs from these points too. These graphs must actually be a sub-graph to the original graph if control points are accurately found. Fig 2 shows the reference DEM in which the control points have been marked manually. Fig 3 shows the DEM of a candidate ASCII file in which control points have been found along with a false-positive matches. Some of the control points have been highlighted.

6. Find the orientation of the sub-graph with respect to the parent graph. Accordingly do the vertical and horizontal orientation of the DEMs. Form a third ASCII file combining the data from the parent DEM and the best matched candidate DEM. (Direct joining of tiles of data is possible by providing latitude and longitude data.)

7. Display the ASCII file by using dynamic colour coding as shown in Fig 4.

The salient points of this technique are that though sub-graph matching is said to be a computationally difficult, but since only a limited number of matching points are considered, it gives a better solution as compared to other feature-point finding techniques. Another advantage of this technique is that it can work with almost any resolution DEM. Any DEM which can be converted to be proper ASCII values can be used as an input. Since the parent ASCII DEM is already subdivided into manageable children ASCII files, the time complexity greatly reduces for the DEM registration.

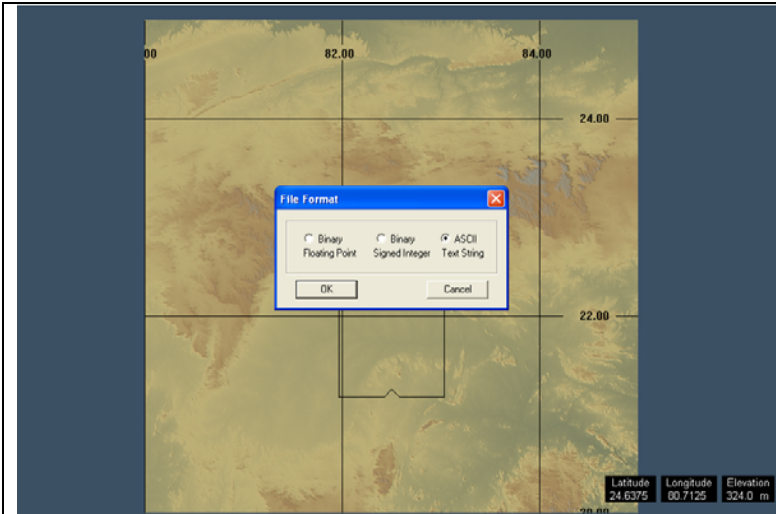

Fig. 1 Convert a DEM from Geotiff format to ASCII file
*Courtesy: http://3dem.ucsd.edu/*

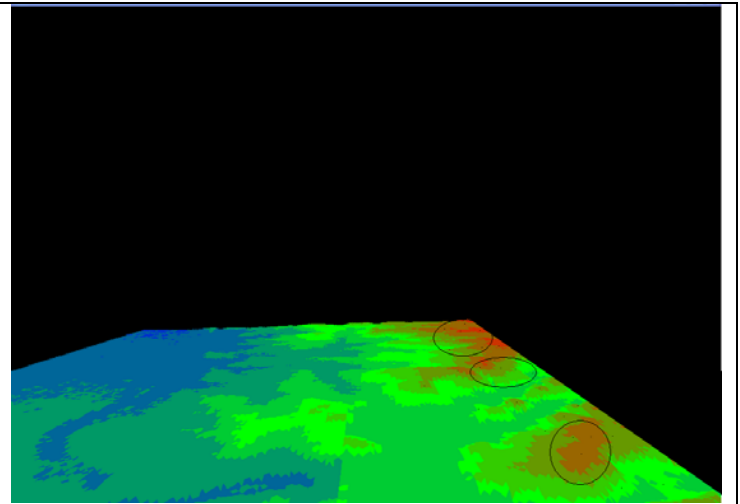

Fig. 2 Control Points marked in the reference DEM as small black dots.

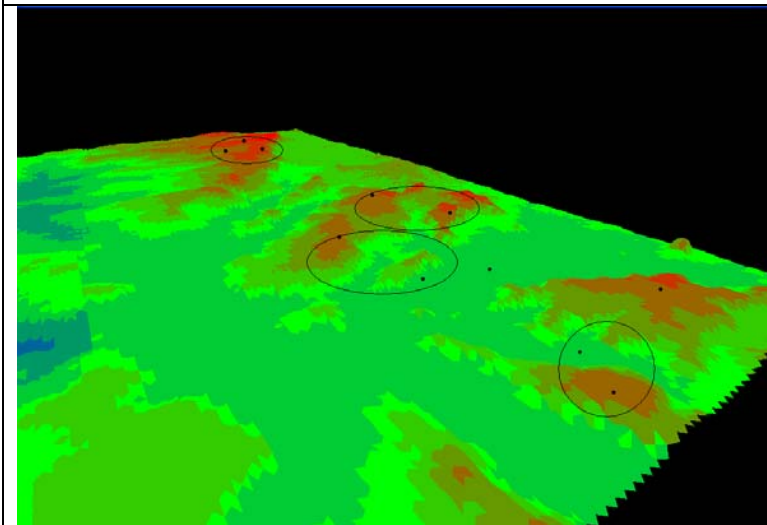

Fig. 3 Control points found in the candidate DEM. Few false-positives are also seen.

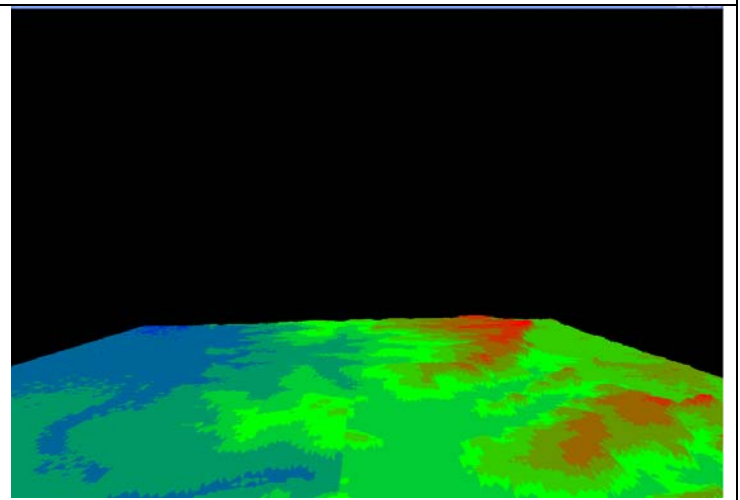

Fig. 4 Display of the registered DEM (registered from reference and candidate DEM).

## IV. Results

For experimentation initially 40 control points were considered from a matrix of 300 x 300 points. These points were marked manually. The various criterias for evaluation of error considered were – mean value comparison, Root Mean Square Error (RMSE) finding, Total Squared Curvature error and t-statistic error analysis.

| Matching technique | Error Analysis (difference in %) | | | |
|---|---|---|---|---|
| | Mean difference | Root Mean Square Error (RMSE) | Total Squared Curvature | t-statistic error analysis |
| Direct Control Point matching | 12.5 | 8.23 | 3.45 | 3.25 |
| Edges matching along congruent quadrangle of the control point. | 8.56 | 7.78 | 3.01 | 3.01 |

Though only 40 points have been considered as of now, some of the common problems faces were – detection of too many false matching due to thresholding, huge margin for error removal and time & complexity heavy computation. Also, presently only same resolution DEMs were used for experimentation. The error factors would increase as the complexity of DEM grows.

## V. Conclusion

The literature describes the experimentations performed for registration of DEMs. For this experimentation DEMS of same resolution have been used. The methods used for registration, has been described. Results in terms of error evaluation have been shown. For display of ASCII data, dynamic colour coding is used which lacks terrain look but still manages to give a good idea of the height difference in the area. The feature-point finding and matching technique may be extended by using morphological operations.